\documentclass[a4paper,fleqn]{cas-dc}

\usepackage{booktabs}
\usepackage{tikz}
\usepackage{inconsolata}
\usepackage{algorithm}
\usepackage{algorithmic}
\usepackage{multirow}
\usepackage{multicol}
\usepackage{quoting}
\usepackage[numbers]{natbib}
\usepackage[format=plain]{caption}
\def\tsc#1{\csdef{#1}{\textsc{\lowercase{#1}}\xspace}}
\tsc{WGM}
\tsc{QE}
\tsc{EP}
\tsc{PMS}
\tsc{BEC}
\tsc{DE}

\begin{document}
\let\WriteBookmarks\relax
\def\floatpagepagefraction{1}
\def\textpagefraction{.001}

\shorttitle{Leveraging social media news}

\shortauthors{CV Radhakrishnan et~al.}

\newcommand{\dataset}{\textsc{WellXplain}}
\title [mode=title]{\textsc{WellXplain}: Wellness Concept Extraction and Classification in Reddit Posts for Mental Health Analysis}




%
\author[1]{Muskan Garg}[]

\cormark[1]

\fnmark[1]

\ead{garg.muskan@mayo.edu}


\credit{Conceptualization of this study, Methodology, Data curation, Writing - Original draft preparation}

\affiliation[1]{organization={Mayo Clinic},
    city={Rochester},
    postcode={55901 MN}, 
    country={USA}}

\begin{abstract}
Amid the ongoing mental health crisis, there is an increasing need to discern possible signs of mental disturbance manifested in social media text. Neglecting multi-dimensional aspects of social and mental well-being (i.e., wellness dimensions) over time can adversely affect an individual's mental health. During in-person therapy sessions, manual efforts are used to identify the causes and consequences of triggering latent factors of mental disturbance, which is a meticulous and time-consuming task for mental health professionals. To enable such fine-grained mental health screening, 
we define the task of determining wellness dimensions in Reddit posts as wellness concept extraction and classification problem. We construct a novel dataset called \dataset, which consists of 3,092 instances and a total of 72,813 words. Our experts developed an annotation scheme and perplexity guidelines for annotation based on a well-adapted Halbert L. Dunn's theory of wellness dimensions. Further, the data encompasses human-annotated text spans as pertinent explanations for decision-making during wellness concept classification. We anticipate that releasing the dataset and evaluating the baselines will facilitate the development of new language models for concept extraction and classification in healthcare domain.
\end{abstract}


\begin{highlights}
\item Introducing the need of datasets for reliable simulations in mental healthcare.
\item Corpus construction for wellness concept extraction and classification.
\item Analyzing domain-specific transformers and large language models for this task.
\item Examining reliability of traditional multi-class classifiers.
\end{highlights}

\begin{keywords}
 Corpus construction \sep mental health \sep WellXplain \sep wellness dimensions
\end{keywords}

\maketitle
\section{Introduction}

A clinically significant impairment in a person's intellect, emotional control, or behavior is what is known as a mental disorder, suggesting cognitive decline. The UN Resolution of ``Transforming our World: the Agenda 2030 for Sustainable Development'' adopted in September 2015~\cite{un2015transforming}, outlined an ambitious vision to tackle \textit{Goal 3: Ensure healthy lives and promote well-being for all at all ages} of Sustainable Development Goals (SDG). By 2030, UN plans to reduce by one-third premature mortality from non-communicable diseases through prevention/treatment and promote the mental health and well-being.\footnote{https://www.un.org/development/desa/disabilities/envision2030-goal3.html} 
Untreated depression is conjectured to be the leading cause of suicide~\cite{garnett2022suicide}. Reports released in August 2021\footnote{https://www.theguardian.com/society/2021/aug/29/strain-on-mental-health-care-leaves-8m-people-without-help-say-nhs-leaders} indicate that \textit{1.6 million people} in England were on waiting lists to seek professional help with mental health care.

Despite significant technological advances, mental health assessment remains a dark mark on public health efforts. Causes for mental health disturbances are broad, including an unspecific gamut of factors such as physical health problems, social conflicts (e.g., bullying, prejudice, stigma, race issues), abuse, grief, financial and professional difficulties, etc.. 
These causes are aggravated when patients do not disclose their concerns to mental health professionals, rather find solace on social media \cite{l2012similarities}. Motivated with non-intrusive high value information, social media data lessens the effect of limited availability of mental health practitioners. In these dire circumstances, online platforms are frequently relied upon not only as open and unobtrusive sources of information, but also as a place for honest disclosure, where people may freely express themselves along with their thoughts, beliefs, and emotions~\cite{resnik2021naturally}.

\begin{figure*}[t]
    \centering
    \includegraphics[width=0.79\textwidth]{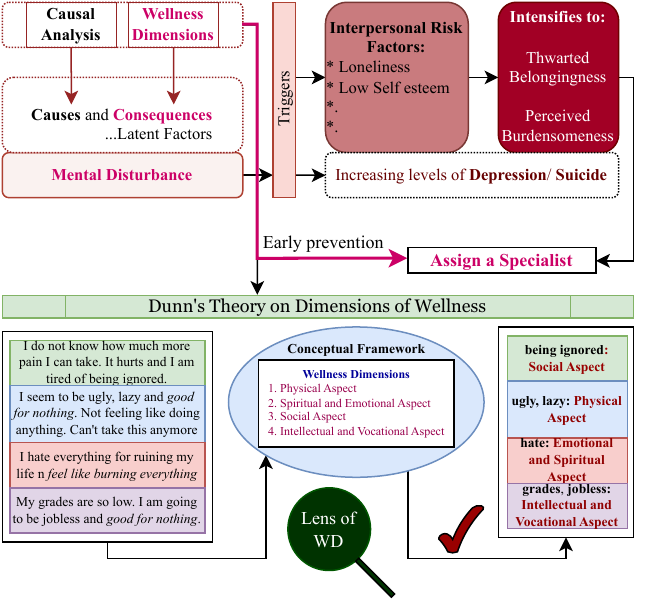}
    \caption{An illustration of the utility of \textsc{WellXplain}. The intrinsic auxiliary tasks of wellness identification influence AI model's focus. We simulate the process of the early wellness concept extraction that undergo initial screening of a given Reddit post to determine the wellness concepts impacting mental health. The wellness dimensions in our work are grounded through Dunn's theory of wellness.}
    \label{fig:overview}
\end{figure*}

However, social media is extremely noisy because of popular culture references and slang terms that are pervasive in online expressions. This noise makes it hard to develop automated methods for mental health screening methods that levels with mental health professionals \cite{berryman2018social,pourmand2019social}. Furthermore, prior work on mental health analyses from social media focuses on assessing posts that already exhibit particular mental health traits (e.g., analyzing mental health subreddits related to suicidality and depression) \cite{o2017linguistic,franco2018systematic,du2018extracting}. 

Amid the huge social impact of COVID-19 pandemic, the research community witness the presence of mental disorders in an individual through consequential affect on wellness dimensions due to prevailing reasons behind mental disturbance. We do not intend to invalidate the prior works with causal analysis such as CAMS, but expect to support the existing works with clinical concept extraction through our newly proposed wellness dimensions dataset \textsc{WellXplain}. We formulate this problem as a multivariate task to construct and release a dataset to develop a comprehensive and contextual AI models that determines all the wellness dimensions that are being affected, in a given Reddit post.


\noindent\paragraph{Social Media.} We focus on Reddit, as it appeals to a significant number of subscribers due to its anonymity feature. For instance, \emph{r/depression} subreddits has 934K and  \emph{r/suicidewatch} has 401K subscribers. These communities have enabled individuals to express their personal experiences and seek support for their mental conditions. The online users may suffer mental disturbances long before explicitly writing about it. Such individuals may be helped through early detection of negative mental health outcomes and subsequent intervention. This critical situation underscores the need for techniques to identify potential wellness concerns as clinical concepts in self-narrated texts submitted online, so as to be able to point out pertinent sources of professional counsel and therapy. 

Motivated by the need for automation in mental healthcare over social media resource points, we expand existing AI-driven models towards clinical concept extraction through wellness dimensions in human writings. Without such systems, a cry for help may remain unnoticed and ignored, or it may receive an overly generic message rather than one that addresses the specific mental health aspect. However, recent studies suggest that with the increasing deployment of AI-driven models on social media data, there is a growing risk of people turning to them as an outlet for disclosing personal misfortune~\cite{harrigian2021state,roy2021machine}. To this end, we further incorporate the clinical settings for clinical concepts extraction in self-narrated texts on Reddit.

%



%

Due to increased demand for quality mental healthcare and limited availability of mental health professionals, there exists a need to simulate the practice of identifying causes and consequences behind mental disturbance. Prior work has focused on coarse-granular mental health classification, which ignore a diverse array of underlying issues\cite{choudhury2010global, gaur2019knowledge}. According to the survey, individuals who lack strong familial and friendship ties are ten times more prone to mental health challenges\footnote{\url{https://mentalstateoftheworld.report/}}. The Interpersonal-Psychological Theory of Suicide~\cite{joiner2009interpersonal} states that a serious suicidal behavior is rooted in the combination of \textit{acquired capability}, \textit{Perceived Burdensomeness} (\textsc{PBu}: hallucination of being disconnected/ burden on society), and \textit{Thwarted Belongingness} (\textsc{TBe}: feeling of being isolated), which are collectively necessary and sufficient proximal causes evolving from affected Wellness Dimensions (see Figure \ref{fig:overview}). As such, we choose to simulate the problem of identifying clinical concepts in the form of wellness dimensions for Reddit posts reflecting suffering and distress. This task is well rooted in multiple clinically-grounded applications including:
\begin{enumerate}
    \item \textit{Fine-grained Natural Language Understanding}: A fine-grained NLU can discern wellness from user inputs, like identifying social aspect from not just explicit statements ("I am lonely") but also implicit ones ("I don't feel like getting out of bed but want to make friends"). It can understand the depth and subtleties in a user's language, such as differentiating between transient moods and persistent wellness states or detecting masked concepts behind seemingly deteriorating mental health. The model would also consider the context in which certain phrases are used to represent the wellness dimension, ensuring that it doesn't misinterpret casual uses of terms that might have psychological implications. With the capability to understand nuances, the system becomes more sensitive to users' psychological states, potentially identifying issues before they escalate. The deep understanding facilitates a more individualized therapeutic experience, accounting for the unique ways in which people express mindfulness.
    \item \textit{Clinical Concept Extraction}: Clinical concept extraction leverages NLP methods to autonomously detect and pull medically pertinent details from non-structured textual data. Algorithms for extraction can be tailored to pinpoint and highlight particular keywords or expressions, especially those indicating the reasons (wellness dimensions) leading to outcomes (cognitive decline) like "lonely", "headache", or "poor grades". Automated platforms can alert human oversight teams or moderators about posts suggesting significant risk, ensuring faster response. By discerning dominant topics or challenges in such communities, moderators can compile resources, posts, or actions that cater to these specific needs. 
    \item \textit{Easy-to-reach-out 'counselors' through early detection of cognitive decline}: Many individuals may not always recognize the need to reach out or might hesitate to approach professionals. Thus, having an automated system that identifies potential wellness issues in their social media posts can bridge this gap and offer timely assistance. The AI-driven model shall detect keywords, phrases, patterns, or sentiments that suggest a potential wellness dimension that needs attention. Rather than waiting for someone to acknowledge and vocalize their issues, the system identifies potential problems and suggests intervention. Addressing wellness concerns at an early stage can lead to better outcomes and prevent issues from escalating. The automated nature can provide a non-judgmental space for users to express their feelings. AI-driven models can learn and adapt based on individual user data, offering more personalized guidance over time.
    \item \textit{Social Determinants of Health}: Social Determinants of Health (SDOH) refer to the conditions in which people are born, grow, live, work, and age, including the wider set of forces and systems shaping the conditions of daily life. When applied to subreddits such as r/depression and r/suicidewatch, the SDOH can help in understanding the underlying societal and environmental factors contributing to mental health and wellness. The SDOH provide a lens to understand the external factors that influence the multifaceted nature of wellness. While many users may post about their feelings or emotional states, digging deeper to understand the societal influences on these feelings can lead to more holistic support and interventions. For instance, Experiencing discrimination, social isolation, or a lack of social support can profoundly affect emotional well-being. Positive community ties, on the other hand, can enhance emotional resilience.
    
\end{enumerate}

\noindent \paragraph{Contribution.} We introduce an AI-based system for wellness dimensions that uses techniques to identify the concerns affecting wellness dimensions that may require counseling and therapy. The goal of such a system is to provide an efficient and effective way to identify individuals who may be struggling with wellness dimensions and to guide them towards appropriate support and resources. A recent surge in quantifying wellness dimensions suggest low-level analysis of personal writings for depressive symptoms in lieu of any of the six dimensions of wellness: 
different dimensions of wellness as:
(i) physical aspect, (ii) intellectual and vocational aspect, (iii) social aspect, (iv) emotional and spiritual aspect. Such stressful situational aspects, prevailing for a long duration, may result in depressive symptoms. To this end, we consider the affected wellness dimensions (situational effects) as a major consequence of mental disturbance and propose the analysis of \textit{wellness dimensions} grounded in established theory of High-level Wellness by Halbert L.\ Dunn \cite{dunn1959high}, so as to investigate the evolution of mental health disturbances in social media posts. According to Dunn, high-level wellness is a state of complete physical, mental, and social well-being, beyond just the absence of disease or infirmity. He argued that high-level wellness is a dynamic process that requires ongoing effort and attention to maintain. Consider the following example:
\begin{quoting}[]
        From dealing with the fallout of my ex, to stressors at work, nothing compares to true \textcolor{blue}{loss of my baby boy} $\leftarrow$ \textcolor{red}{(Social Aspect)}. To everyone feeling shitty this New Year’s Eve, you are not alone.\\
\end{quoting}

\noindent Here, the author expresses three different situations, including problems with relationship status (Social Aspect), problems with work and career (Intellectual and Vocational Aspect), and loss of their beloved child (Social Aspect). However, the major focus of this text is the \textit{loss of a baby boy}, emphasizing a major impact of life circumstances on the \textit{social aspect}. We shall discuss such perplexities in more detail further below.
These may pertain to a number of \emph{dimensions of wellness} studied in psychology. For, to truly address and attenuate mental disturbances, we need datasets that can bridge the gap between life's challenges and potential mental health disorders. This will expedite early detection and intervention. By understanding and identifying these wellness concerns as clinical concepts, we can pave the way for more accessible counseling resources, offering immediate support to those grappling with distress. As such, our task suggests the new research direction of extracting cause-and-consequence in a given text where cause (wellness concept extraction) aids in early detection of consequences (cognitive decline/ mental health illness).

\begin{table*}
\centering
\begin{tabular}{p{2.5cm}p{13cm}p{0.5cm}}
\toprule[1.5pt]
\textbf{Dataset} & \textbf{Details}   & \textbf{Avail.} \\
\toprule[1.5pt]

\textbf{CLPsych}~\cite{coppersmith2015clpsych} & Three types of annotated information using Depression, Control, and PTSD & S \\
\hline

\textbf{MDDL}~~\cite{shen2017depression} & 300 million users and 10 billion tweets in Depression, Non-depression, and Depression candidate  & A \\
\hline

\textbf{RSDD}~\cite{yates2017depression} & Reddit dataset of 9210 users in depression and 1,07,274 users in control group & ASA \\
\hline

\textbf{SMHD}~\cite{cohan2018smhd} & Reddit dataset for multi-task mental health illness  & ASA \\
\hline
\textbf{eRISK}~\cite{losada2018overview} & Early risk detection by  CLEF lab about problems of detecting depression, anorexia and self-harm & A \\

\hline



\textbf{Sina Weibo}~\cite{cao2019latent}  & 3,652 (3,677) users with (without) suicide risk from Sina Weibo &  AR \\
\hline
\textbf{SRAR}~\cite{gaur2019knowledge} & Posts from 500 Redditors (anonymized) \& annotated by domain expert &  ASA \\
\hline
\textbf{Dreaddit}~\cite{turcan2019dreaddit} & 190K Reddit posts of 5 different categories &  A \\
\hline
\textbf{GoEmotion}~\cite{demszky2020goemotions}  & Manually annotated 58k Reddit comments for 27 emotion categories & A \\
\hline
\textbf{UMD-RD}\footnote{The University of Maryland Reddit Dataset}~\cite{shing2020prioritization} & 11,129 users who posted on r/SuicideWatch and 11,129 users who did not & ASA \\
\hline
\textbf{SDCNL}~\cite{haque2021deep} & Reddit dataset of 1895 posts of depression and suicide  & A \\
\hline
\textbf{CAMS}~\cite{garg2022cams} & Interpretable causal analysis of mental illness in social media (Reddit) posts & A \\
\bottomrule[1pt]
\textsc{\textbf{WellXplain}} & Wellness concept extraction in Reddit posts & A \\ 
\bottomrule[1.5pt]
\end{tabular}
\caption{Different mental health datasets and their availability. A: Available, ASA: Available via Signed Agreement, AR: Available on Request for research work}
\label{tab:dataset_ex}
\end{table*}


    
      
        
        
         
    

While there are publicly available textual mental health assessment datasets~\cite{garg2023mental}, enabling AI models to account for wellness dimensions may yield rich and fine-grained contextual information, ultimately also relevant to clinicians (see Table~\ref{tab:dataset_ex}). To this end, we observe a scarcity of available datasets for the finer-grained task of identifying clinical concepts through wellness dimensions as a consequence of poor well-being. To the best of our knowledge, existing literature has no language resources on wellness dimensions that are available for research and development. The reason behind limited public availability is due to (i) limited low-level analysis for cause-and-consequence of mental health and (ii) sensitive nature of the data. 
The public release of our dataset on GitHub\footnote{\url{https://github.com/drmuskangarg/WellnessDimensions/}} shall facilitate future research by raising the red alert on determining one of the pre-defined concepts (wellness dimensions) affecting mental disturbance via a 
longitudinal study of Reddit posts~\cite{tsakalidis2022identifying}.
Thus, our \emph{key contributions} can be summarized as follows:
\begin{enumerate}
    \item To the best of our knowledge, this is the pioneer quantitative investigation emphasizing the importance of concept extraction using wellness dimensions to analyze mental health in Reddit posts.
    \item We construct and release \textsc{WellXplain}, a new English dataset of 3,092 instances with 72,813 words as a \textit{multi-class classification} problem with \textit{text-span} explanations. 
\item Our experiments with machine learning classifiers, sequence-to-sequence models, Transformer models and large language models set the baseline and suggest the opportunities for further research and development.
\end{enumerate}  

\section{Corpus Construction}
The current landscape offers abundant open-source datasets for research and application development across various fields. However, the scarcity of high-quality datasets is a pressing issue, particularly in the rapidly evolving field of Natural Language Processing (NLP), where they are vital for training and evaluating Language Models. 

Real-world datasets are often complex, unorganized, and lack structure, and the performance of models is closely tied to the dataset’s quantity, quality, and relevance. It’s crucial to understand a dataset's nature, significance, and its foundational role in building artificial intelligence-based NLP systems. Bearing this in mind, the principal obstacle in dataset creation lies in safeguarding the uniformity and trustworthiness of the annotations. To surmount this challenge, we construct the psychology-driven annotation scheme using Dunn's theory of wellness dimensions and experts' intervention for annotating dataset and deploy it. In this section we first discuss the data acquisition, followed by annotation scheme design and annotation task. 

\subsection{Data Acquisition}
Reddit is a social media platform for open discussions and its ability for individuals to post anonymously make it a diverse platform for candid and personal data collection about mental health complexities based on personal experiences. Being anonymous, people express their thoughts and share their experiences about mental disturbance with ease, often facilitated by a non-judgmental environment as compared to other prominent social media platforms. Although the Python Reddit API Wrapper (PRAW) API, an interface used for data collection from Reddit, allows the retrieval of information about authors and creation date, we limit the information to the title and text of the post to maintain the privacy of data. The Reddit platform provides an opportunity of sharing \textit{user-generated} and \textit{user-curated} context in the community-specific \textit{subreddits} for open discussions indicating special themes or conditions. In the past five years, the subreddits associated with mental disorders such as \texttt{r/depression} and \texttt{r/SuicideWatch} have grown more than 100\% and 250\%, respectively. A limit on the number of characters is 10,000 for a comment, 40,000 for a post, and 300 for post titles.
Prior work on mental health analysis in Reddit posts suggests the use of next word prediction models, which are often computationally expensive for lengthy posts. Although we suggest the need of a more comprehensive and contextual approach to solve the problem, we keep the maximum length of the Reddit posts at 256 tokens in our dataset to facilitate machine learning research. The use of NLP for Reddit posts has major challenges of double negation, syntactic and semantic ambiguity, as well as commonsense and domain specific knowledge infusion.

To limit the length of a Reddit post, we limit the maximum number of words in each post to 256. We performed expert-guided manual cleaning and filtered the Reddit posts to limit them to \textit{user experience} impacting mental disturbance, resulting in 3,092 posts. Our three experts: a senior clinical psychologist, a rehabilitation counselor, and a social NLP researcher, frame annotation schemes and perplexity guidelines to annotate a given Reddit post with one of the pre-defined dimensions of wellness.

\subsection{Annotation Scheme Design}

\noindent \textbf{Wellness Dimensions.}
Recent literature suggests six different dimensions of wellness as a consequence of mental and social well-being, namely, (i) physical aspect, (ii) intellectual aspect, (iii) vocational aspect, (iv) social aspect, (v) emotional aspect, and (vi) spiritual aspect \cite{kitko2001dimensions,strout2012six}. Our experts performed a pilot study on $40$ instances, observing a high overlap and ambiguity of words and interpretations, respectively, in two different cases, (i) intellectual vs.\ vocational aspect, (ii) emotional vs.\ social aspect. The observations are supported by Dunn \cite{dunn1959high} in his model of \textit{Wellness as a Conceptual Framework}, which defines wellness as an integrated balance of social, physical, cognitive (intellectual and vocational), and spiritual (emotional) health.

We further investigate the corresponding consequences of these cause on similar lines through the psychologically-driven theory of \textit{dimensions of wellness} by Dunn, and define them as:
\begin{itemize}
    \item \textbf{Physical Aspect (PA)}: Physical development encourages a good diet and nutrition, while discouraging the use of tobacco, drugs, and excessive alcohol consumption. Optimal wellness is met through good exercise, good sleep, energetic behaviour, enthusiasm, and eating habits. Body shaming affects the physical well-being of a person, making them realize their problems with medical history and physical appearance.
    \item \textbf{Intellectual and Vocational Aspect (IVA)}: The use of intellectual and cultural activities within or beyond the classrooms, combined with the human and learning resources. Wellness of a person cherishes intellectual growth and stimulation. The occupational dimension recognizes personal satisfaction and enrichment in one’s life through work. It affects their attitude towards creative thinking, professional growth and other financial expenses.
    \item \textbf{Social Aspect (SA)}: The social dimension emphasizes the interdependence between society and nature. A person becomes more aware of their importance in society as well as the impact which they may have on multiple environments. Social connections help in flourishing interpersonal factors of their nature and enhance the ability to emphasise cultural impacts.
    \item \textbf{Spiritual and Emotional Aspect (SEA)}: The spiritual dimension recognizes the meaning and purpose in human existence. It includes the development of a deep appreciation for the depth and expanse of life and natural forces that exist in the universe. It is characterized by a peaceful harmony between internal personal feelings and emotions. The emotional dimension enhance the awareness and acceptance of one’s feelings and the extent of positivism and enthusiasm about one’s self and life. It includes the capacity to manage one’s feelings and related behaviors including the realistic assessment of one’s limitations, development of autonomy, and ability to cope effectively with stress.
\end{itemize}

\begin{table*}[!t]

    \centering
    \caption{A sample table to present dimensions of wellness and annotated samples. Here, column 1 presents four WD: Wellness Dimensions, namely, PA: Physical Aspect, IVA: Intellectual and Vocational Aspect, SA: Social Aspect, and SEA: Spiritual and Emotional Aspect.}
    \begin{tabular}{p{1cm}|p{10cm}|p{3.5cm}}
        \midrule \midrule
        \textbf{WD} & \textbf{Annotated samples} & \textbf{Text Span}\\
        \midrule \midrule
        \multirow{6}{*}{PA} & My stomach bulges out too much, my face is fat, my acne is ugly, I need to shave, my teeth are yellow, my butts too small, etc & face is fat, acne is ugly, teeth are yellow \\
         & \text{-----------------------------------------------------------------------------------------------------------------------------------} \\
        & Well it did, to the point where any pressure on my stomach caused me pain and I would have to just lay down until the upset stomach subsided & upset stomach\\
        \midrule
        \multirow{4}{*}{IVA} & I've almost failed 3 classes the past 3 school years, and I'm on the brink of failing another one & failed 3 classes the past 3 school years\\
         & \text{-----------------------------------------------------------------------------------------------------------------------------------} \\
        & I'm a 23 year old unemployed woman still living with my mom & unemployed woman \\
        \midrule
        \multirow{7}{*}{SA} & Bad family, just got out of a bad relationship, have always been bad at making friends & bad at making friends \\
        & \text{-----------------------------------------------------------------------------------------------------------------------------------} \\
        & I'm 21, both parents dead, no family support, just friends, my brother abandoned me and left me out in the cold and kicked me out and cut me off because I was out drinking as a teenager & both parents dead, no family support, just friends, my brother abandoned me \\
        \midrule
        \multirow{6}{*}{SEA} & Failing half my classes, same as last semester, going to drop out of college at the end of the semester with no life direction, no ambition, no motivation, no desires, dreams, will to live & no ambition, no motivation, no desires, dreams, will to live\\
         & \text{-----------------------------------------------------------------------------------------------------------------------------------} \\
        & I want someone to hold me and prove me wrong and tell me that I am valued and not worthless & worthless \\
        \midrule \midrule
    \end{tabular}
    \label{tab:samples}
\end{table*}
\noindent This list is not exhaustive, but is a starting point for our study, giving rise to a final set of four dimensions, namely, (i) Physical Aspect (\textsc{pa}), (ii) Intellectual and Vocational Aspect (\textsc{iva}), (iii) Social Aspect (\textsc{sa}), (iv) Spiritual and Emotional Aspect (\textsc{sea}). 
Samples from the dataset are given in Table~\ref{tab:samples}.

\paragraph{Guideline development.} Our experts developed annotation guidelines by using the definitions of Wellness Dimensions given by Dunn \cite{dunn1959high} as mentioned above. Given that wellness dimensions are a highly subjective and complex issue, identifying them accurately can be challenging and prone to errors with naive judgment. To overcome this, our team of experts negotiates a trade-off between \textit{using text-based marking} for developing advanced AI models and \textit{reading between the lines} to provide psychological insights, while framing annotation schemes. Our experts developed annotation guidelines to aim at:

\begin{enumerate}
    \item \textit{Finding potential text spans to identify dimensions of wellness affecting mental health.}
    \item \textit{Developing a language resource to build explainable AI models for (content-aware domain-specific) 4-class classification problem of determining dimensions of wellness.}
\end{enumerate}


The experts annotated 40 samples of the dataset in isolation, using fine-grained guidelines to avoid biases \cite{mitra2015credbank}. Annotations are crucial in supervised machine learning, where you train a model based on labeled examples. 'Fine-grained' suggests that these rules are very specific, aiming to capture nuances in the data. Halbert L. Dunn's concept of "High-Level Wellness" in 1959 is among the foundational works on the modern understanding of wellness. Dunn proposed that wellness is not merely the absence of disease but a proactive, holistic approach to life that combines the physical, mental, and social aspects, among others. By using Dunn's model of Wellness as a foundation for creating annotation rules, the project anchors its methodology in a well-established, researched framework. This can lend credibility to the endeavor, as it's not based on arbitrary or ad hoc criteria but on a historical and scholarly model. Using fine-grained rules means that the project is making efforts to capture the subtleties and nuances of wellness as described by Dunn. For instance, instead of broadly labeling a piece of data as 'wellness-related', there might be more specific labels like 'emotional wellness', 'physical wellness', 'social wellness', etc., all derived from Dunn's conceptualizations. One of the biggest challenges in AI and machine learning is ensuring that models do not inherit biases present in data or human annotators. By relying on a structured, well-defined framework for annotations, the aim is to provide a standardized basis for labeling data, minimizing subjective biases that could arise from individual interpretations. It's essentially a way to systematize and standardize the process, ensuring that different annotators would (ideally) label the same piece of data in the same way. A machine learning model's predictions are as good as the data it's trained on. If the training data is annotated with consistency, precision, and alignment with a recognized framework like Dunn's, the resulting model is more likely to make accurate and reliable predictions in line with that framework.

Therefore, using Dunn's model-based fine-grained rules for annotations helps ensure that the annotations are more objective and consistent, leading to more reliable and trustworthy results. The current annotation scheme has limitations in capturing all the aspects of the phenomenon. Despite these efforts, due to the subjective nature of this complex task, we encountered possible dilemmas. Therefore, we propose a set of perplexity guidelines that are intended to simplify the task and make future annotations easier to perform.


\subsection{Perplexity guidelines} The task of annotating text for wellness dimensions can be complex, especially when the text mentions multiple reasons or contains ambiguity in its interpretation. In order to simplify the task and facilitate future annotations, we have developed a set of perplexity guidelines. These provide a framework for annotators to follow, ensuring that their annotations are consistent and accurate, thereby reducing the potential for errors or misunderstandings when interpreting the text. By using these guidelines, we made the task of annotating wellness dimensions more manageable and accessible, ultimately leading to a better understanding of the relationship between text and well-being. We enumerate the major perplexity guidelines as follows:
\begin{enumerate}
    \item \textbf{Presence of Multiple Aspects}: Our team of experts has encountered posts on social media platforms where individuals express their feelings due to multiple reasons. To overcome this challenge, our team suggests a solution where the specific text spans in the post that contribute to the more focused health consequences should be identified and associated with the corresponding wellness dimension. This approach allows for a more nuanced annotation of the wellness dimensions and provides a clearer understanding of the underlying causes of mental health issues. Consider the following post $P_1$:
    \begin{quote}
        $P_1$: I cannot do anything without screwing it up, I just got suspended from school my family, my friends, all lose their trust in me, I'm just not cut out for anything I don't think the world has a place for me.
    \end{quote}
    In the given post $P_1$, suspension from school affects the IVA of the author, resulting in disturbed SA. This situational effect on wellness dimensions results in mental disturbance. For instance, in a given post $P_1$, the affected dimension of wellness is SA.
    
    \item \textbf{Annotation Ambiguity}: As wellness dimensions are not atomic in nature, identifying them is a highly subjective and complex task, which can be challenging for naive human annotators. Even experts may provide varying annotations for the same post, adding to the difficulty of the task. For instance, consider a post $P_2$:

    \begin{quote}
        $P_2$: I hate my home I hate my family and I hate my life I normally can ignore it but sometimes it just gets too much and I end up like I am now crying at my desk with no real reason why
    \end{quote}
    Although there are multiple aspects in this post $P_2$, we must consider the holistic aspect as per the experts' opinion. For instance, the above post is affecting \textsc{sea}, as the user mentions their feelings about lifestyle. 
    
    \item \textbf{Reading between the lines}: The text may contain implicit or subtle hints that suggest a particular wellness dimension, which can be difficult for annotators to identify. We follow a strict guideline for text-span extraction-based classification and avoid making assumptions about the narrator. However, our team of experts has agreed that clear and meaningful words that suggest one of the four wellness dimensions should be annotated accordingly. This approach ensures that the annotations are accurate and consistent, even in cases where the text may not explicitly mention a particular wellness dimension. For instance, consider the following post $P_3$:

    \begin{quote}
        $P_3$: I think being bored 9-5 is even more depressing! Now I'm desperately struggling to keep happy and I have this low to medium level of anxiety that just won't go away no matter how hard I try and think differently.
    \end{quote}
    The given post $P_3$ clearly mentions \textit{9-5} which, as per commonsense, is interpreted as referring to working hours without the need for further assumptions. Thus, the monotonous work environment leads to an annotation of this post as \textsc{iva}. 
\end{enumerate}

\subsection{Annotation task.} To ensure consistency among annotators, the annotation and perplexity guidelines are used for providing formal training to all three student annotators. Our team trains three postgraduate students for manual annotations. Through three successive sessions of trial and error analysis, which involve the annotation of 40 samples per session, we aimed to achieve coherence among the annotators. After the training sessions, we asked the students to perform a two-fold annotation of each data point: (i) Assigning a Wellness label, and (ii) Identifying the text spans that explain the reason for the given label.

\paragraph{Inter-annotator agreement.} 


In our endeavor to recognize wellness dimensions from textual data, we have delineated two specific annotation tasks:

\begin{enumerate}
    \item \textbf{4-Class Classification Task}: The primary objective here is to classify a provided text into one of the four predetermined wellness dimensions. Given the nature of our dataset, which consists of relatively short data points, our initial approach involves multi-class classification. This not only aids in pinpointing the most influenced wellness dimension by the given text but also facilitates an assessment of the robustness and statistical properties of our dataset.

    \item \textbf{Text-Spans Identification}: This task zeroes in on identifying specific segments or spans of text that serve as indicative markers for the wellness dimensions. These spans provide a granular understanding, shedding light on which particular sections of the text resonate with the wellness dimension of interest.
\end{enumerate}

To ensure the \textit{reliability} of our annotations, particularly for the 4-class classification, we employed multiple validation measures. Initially, three independent student annotators were engaged in this task. Their annotations underwent a statistical evaluation using Fleiss' Kappa inter-observer agreement, yielding a value of \( \kappa = 74.39\% \). This value underscores a significant agreement among them, emphasizing the trustworthiness of their annotations.

To augment \textit{accuracy}, we adopted a majority voting mechanism: for any given data point, if a label was endorsed by at least two out of the three annotators, it was considered as the final label. Yet, we introduced another layer of scrutiny. Recognizing the nuances and intricacies of the domain, these annotations were further vetted by a seasoned clinical psychologist. This ensured that our labels resonated with domain-specific expertise.

For a comprehensive perspective on the data distribution across the distinct wellness dimensions, we compiled a table detailing the statistics. This table, designated as \textsc{WellXplain}, offers insights into the count of data points under each dimension and other relevant details (See Table \ref{tab:2}).

\begin{table}
    \centering
    \caption{Statistics of \textsc{WellXplain}
    }
    \begin{tabular}{lr}
        \toprule[1.5pt]
        \textbf{Criteria} & \textbf{Frequency} \\
        \midrule[1pt]
        \textbf{Statistics}\\
        Number of Posts & 3092\\
        Total number of Words & 72,813\\
        Max.\ number of Words & 231 \\
        Total number of Sentence & 3,376 \\
        Max. number of Sentences/post & 7 \\
        \midrule
       \textbf{Wellness Dimension}\\
        Physical Aspect & 750\\
        Intellectual and Vocational Aspect & 592 \\
        Social Aspect & 1,139 \\
        Spiritual and Emotional Aspect & 621 \\
 
         
         \bottomrule[1.5pt]
    \end{tabular}
    \label{tab:2}
\end{table}
For the task of text-span extraction, our three annotators, combined as a group, were asked to choose a set of words as text-spans representing explanations for corresponding wellness dimensions. 
The student annotators jointly extracted text spans as a potential \textit{explanation} for each data point to form the final list of annotations for explanation. Our three experts carried out an agreement study over the final lists of explanations generated by a group of three student annotators using \textit{Fleiss' Kappa} statistics~\cite{fleiss1971measuring}, resulting in a $\kappa$ score of 87.32\%. The experts were asked to categorize their annotations as either Agree or Disagree. Although the inter-annotator agreement for the former task is slightly lower due to confusion between \textsc{pa} and \textsc{sea}, we observe a higher agreement for the selection of explanatory text-spans.

\subsection{Additional Data Analysis}

We observe the number of samples for different aspects as: \textsc{pa} (24.26\%), \textsc{iva} (19.15\%), \textsc{sa} (36.84\%), and \textsc{sea} (20.09\%). 
The large number of samples for sentiment analysis (SA) indicates that our society may face more challenges in dealing with issues related to ``near and dear ones" and ``loneliness." These issues are often complex and difficult to handle, and may require more attention and resources to address effectively. The high number of samples in SA suggests that individuals are seeking emotional support and validation through social media platforms, which highlights the importance of addressing mental health concerns in these online spaces.
The lower number of samples for \textsc{iva} suggests that there is strong administrative control and good governance in this domain. This could be attributed to stringent regulations and protocols, or the expertise and experience of the individuals involved. The identification of \textsc{iva} incidents may require specialized knowledge or expertise, which could explain the lower number of samples. Furthermore, the importance of interpersonal relationships outweighs that of intellectual and vocational skills. It is crucial to maintain healthy relationships with others, as they play a significant role in our overall well-being. While vocational and intellectual skills are important, they do not have the same impact on our mental health as our interactions with others. Developing and maintaining positive relationships can lead to a more fulfilling life and better mental health outcomes.

\subsubsection{Frequent words in text spans.}
The \textit{text spans} marked
or the \textit{text-segments} in the original texts serve as explanations for decisions regarding wellness dimension selections. 
The explanatory text-spans are analyzed to identify the most frequent words for each aspect, and the results are tabulated in Table~\ref{tab:frequency}.


\begin{table}
\small
    \centering
    \caption{Frequent words in text spans for each wellness dimension.}
    \begin{tabular}{lp{0.8cm}p{5.8cm}}
        \toprule[1.5pt]
        \textbf{WD} & \textbf{Avg.(W)}& \textbf{Most Frequent Words}\\
        \midrule 
        \multirow{2}{*}{PA} & \multirow{2}{*}{4.26} &ugly, anxiety, sleep, meds, pain, tired, panic, drunk, alcohol, diagnosed \\
        \midrule
        \multirow{2}{*}{IVA} & \multirow{2}{*}{5.46} & job, school, work, college, money, failing, failed, time, life, year \\
        \midrule
        \multirow{2}{*}{SA} & \multirow{2}{*}{4.86} & friends, alone, family, lonely, people, feel, want, someone, parents, friend \\
        \midrule
        \multirow{2}{*}{SEA} & \multirow{2}{*}{3.46} & hate, feel, sad, worthless, motivation, anxiety, life, cry, shit, useless \\
    
         \bottomrule[1.5pt]
    \end{tabular}
    \label{tab:frequency}
\end{table}

The data consists of segments of text termed as `text-spans', which likely provide explanations or context about a topic. Analyzing these text-spans is a methodological approach to understand word patterns, occurrences, and the contexts in which words appear.
\begin{enumerate}

\item{Tabulated Results:} The results of this analysis, particularly the most frequently occurring words for each aspect (See Table~\ref{tab:frequency}) offers a breakdown of which words are the most predominantly associated with each wellness concept.

\item {Unique Linguistic Signatures:} Upon careful examination of the table, a noteworthy observation is that the lists of frequent words for each aspect, such as \textsc{sa} and \textsc{sea}, are distinctly separate. This delineation suggests that unique linguistic signatures or patterns are associated with different wellness concepts.

\item {Common Words Across Aspects:} Interestingly, despite the uniqueness in word lists for each aspect, certain commonalities exist. For instance, the word ``feel'' finds its presence in multiple aspects. This revelation emphasizes that some words, despite their universality, are not exclusive indicators of a specific aspect.

\item{Significance of Context:} As exemplars, consider the phrases ``feeling lonely'' and ``feeling useless''. The former is categorized under the \textsc{sa} aspect, whereas the latter falls under \textsc{sea}. Such examples accentuate the pivotal role context plays in language. The meaning and the ultimate categorization of a word can metamorphose based on its adjacent words or the overarching sentiment of the sentence. 
     
\end{enumerate}
The multifaceted nature of language emerges not just from individual words but their composite structures and contextual usage. The dynamic interplay of words in varied contexts poses inherent challenges for AI models, especially those striving for linguistic understanding or classification. Simple frequency-based paradigms might be inadequate, making a strong case for advanced contextual models. An undue focus on keyword frequency sans context could inadvertently lead to misclassification. This accentuates the need for nuanced analysis techniques.



\section{Our Framework}
\subsection{Problem definitions}
The Concept Extraction Framework (CEF) is meticulously designed to automate the process of extracting salient wellness concepts from expansive textual repositories. It seeks to not only localize specific concepts but also discern the intricate relationships or contextual embeddings in which these concepts reside. The mathematical characterization of this task is the multi-class classification challenge for categorizing a given text into one of the predefined wellness concepts.

\subsection{Mathematical Notations and definitions}

The multi-class classification of text into predefined wellness dimensions has emerged as a pivotal task in the realm of textual analysis. In this section, we introduce a mathematical framework for such multi-class classification. We define a classification function that maps a given text to its corresponding wellness dimension. This work leverages probability distributions to capture the likelihood of a text belonging to a particular wellness dimension. To fine-tune our model, we employ the categorical cross-entropy loss, aiming to minimize it across training instances. Optimization techniques, especially stochastic gradient descent, play a crucial role in updating our model parameters and achieving optimal performance. 

\subsubsection{4-class classification task}
Consider the following mathematical notations for the task of multi-class classification:

\begin{itemize}
    \item \( T \): A given text or document that needs to be categorized. 
    \item \( D \): D is the complete set or list of all possible wellness dimensions that you want to classify the Reddit post into. Thus, the set of predefined wellness dimensions, where \( D = \{d_1, d_2, \dots, d_k\} \) and \( k \) is the total number of wellness dimensions. Here, $k=4$ for four different wellness concepts.
    \item \( f \): A classification function which maps a given text \( T \) to one of the dimensions in \( D \).
\end{itemize}

\textbf{Objective}:
When a user posts something on Reddit (or a specific Reddit community focused on wellness), the goal of the classification model would be to automatically determine which dimension of wellness the post is discussing or relating to. Thus, given a Reddit post \( T \), our multi-class classification task aims to determine the most appropriate wellness dimension \( d_i \) from \( D \) to which the text belongs.
\begin{equation}
    f: T \rightarrow D
\end{equation}
\begin{equation}
    f(T) = d_i
\end{equation}
where \( d_i \) is the wellness dimension that best represents the text \( T \).

\textbf{Probability Distribution}:
For many modern classification models, especially those based on deep learning, the output is often a probability distribution over the set of wellness concepts. Let \( P \) be the probability distribution vector generated by the model for text \( T \) over the set \( D \), where:
\begin{equation}
    P = \{p_1, p_2, \dots, p_k\}
\end{equation}
Here, \( p_i \) represents the probability that the text \( T \) belongs to the wellness dimension \( d_i \).

The final classification decision can then be made by selecting the dimension with the highest probability:
\begin{equation}
    d^* = \arg\max_{d_i \in D} p_i
\end{equation}
Where \( d^* \) is the most probable wellness dimension for text \( T \).

\textbf{Loss Function}:
In order to train the model, a suitable loss function is required. For multi-class classification problems, the commonly used loss function is the categorical cross-entropy loss:
\begin{equation}
    L(T, d_i) = -\sum_{i=1}^{k} y_i \log(p_i)
\end{equation}
Where:
\begin{itemize}
    \item \( y_i \) is a binary indicator (0 or 1) if dimension \( d_i \) is the correct classification for \( T \).
    \item \( p_i \) is the predicted probability that \( T \) belongs to dimension \( d_i \).
\end{itemize}

\textbf{Training Objective}:
The main objective during training is to minimize the loss function across all training examples. The parameters of the classification function \( f \) are updated iteratively using optimization algorithms, like stochastic gradient descent, to achieve this goal.

\subsubsection{Text-span Identification}

Attention mechanisms in deep learning allow models to focus on specific parts of the input data. For textual data, this essentially means focusing on certain text-spans or words. The attention weight associated with each word or text-span indicates its importance.

Let $ T = \{w_1, w_2, \dots, w_n\} $ be the sequence of words in text $ T $, where $ w_i $ represents the $ i^{th} $ word.

The attention mechanism assigns a weight $ a_i $ to each word $ w_i $ in $ T $. The sequence of attention weights is given by $ A = \{a_1, a_2, \dots, a_n\} $ where:

\begin{equation}
a_{i} = \text{Attention}(w_{i})
\end{equation}

\textbf{Identifying significant text-spans:} In the pursuit of classifying texts into one of the four wellness dimensions, the importance of specific words or tokens in the text cannot be overstated. These words, which significantly influence the decision-making process of our model, form a unique subset of the textual data. The words  $I$ with the highest attention weights that enables decision-making for the task of 4-class classification is defined as:

\begin{equation}
    I = \{ w_i \mid a_i > \theta, \forall i \}
\end{equation}

where $ \theta $ is a predefined threshold determined empirically in language models.

\textbf{Comparison with Ground Truth:} Let $ G = \{g_1, g_2, \dots, g_m\} $ be the set of text-spans or words that represent the ground truth for the text-spans determining wellness concept in a given text $ T$. The reliability $ R $ of the proposed model can be determined by comparing the set $I$ with $G$:

\begin{equation}
    R = \frac{|I \cap G|}{|G|}
\end{equation}

Here, $ |I \cap G| $ represents the number of words that are both identified as important by the attention mechanism and present in the ground truth. $ |G| $ is the total number of words in the ground truth. The reliability $ R $ gives a ratio of correctly identified important words to the total important words in the ground truth.

\subsection{Large language models}
GPT-3 is a state-of-the-art deep learning model designed for natural language processing tasks. It is the fourth iteration of the GPT architecture and has been trained on vast amounts of text data, making it highly capable of understanding and generating human-like text. The four model variants differ primarily in size and computational demand: (i) \textit{Ada}: The smallest variant, suitable for tasks with limited computational resources, (ii) \textit{Babbage}: Medium-sized, balancing computational demand and performance, (iii) \textit{Curie}: Larger than Babbage, ideal for complex tasks needing higher accuracy, (iv) \textit{Davinci}: The most powerful variant, used for highly demanding tasks but requires the most computational resources. All variants process a given Reddit post to predict a probability distribution over the wellness dimensions. The core architecture comprises multiple transformer layers, with the output being classified into the desired wellness dimensions. Let:
\begin{enumerate}
    \item \( T \) denote a given Reddit post.
    \item \( D \) be the set of pre-defined wellness dimensions, where \( D = \{d_1, d_2, d_3, d_4\} \).
    \item \( M \) represent the set of GPT-3 variants used for classification, given by \( M = \{\text{Ada, Babbage, Curie, Davinci}\} \).
\end{enumerate}

For a given \( T \), a model variant \( m \in M \) predicts a probability distribution \( P_m \) over the set \( D \). Here, \( P_m = \{p_1, p_2, p_3, p_4\} \), where \( p_i \) indicates the probability that the text \( T \) belongs to the wellness dimension \( d_i \). The classification decision for a model variant \( m \) is then:
\begin{equation}
    d^*_m = \arg\max_{d_i \in D} p_i
\end{equation}

Where \( d^*_m \) is the most probable wellness dimension for text \( T \) as predicted by model variant \( m \).

By empirically testing all four variants, the objective is to gauge their performance in classifying Reddit posts into the pre-defined wellness dimensions. This would provide insights into which variant offers the best balance between computational cost and classification accuracy for this specific task.

\subsection{Working Instance}
One of the remarkable features of Transformer-based models like GPT-3 is their attention mechanism, specifically the self-attention mechanism. This allows the model to focus on different parts (text-spans) of an input sentence when producing an output. In the context of our classification task, this mechanism enables the model to weigh the importance of different text-spans in a Reddit post when deciding on the most appropriate wellness dimension.


\begin{enumerate}
    \item \textbf{Input:} \\
    A sample Reddit post, e.g., ``Lately, I've been feeling a bit isolated from everyone. The loneliness is truly setting in."

    \item \textbf{Processing with Attention Mechanism:}
    \begin{itemize}
        \item The model interprets the post using multiple attention heads. Each head concentrates on different text-spans.
        \item One attention head might emphasize the terms ``feeling" and ``loneliness", identifying them as signs of an social state.
        \item Another might give attention to the segment ``isolated from everyone", providing emotional context in the post.
    \end{itemize}

    \item \textbf{Output Probability Distribution:} \\
    The collective understanding from all attention heads results in a probability distribution over the wellness dimensions. The predicted distribution might be:
    
        \text{PA}: 0.08 \\
        \text{IVA}: 0.05 \\
        \text{SA}: 0.75 \\
        \text{SEA}: 0.12\\
    The model, in this case, asserts with a high likelihood that the post is indicative of an 'Social Aspect'.

    \item \textbf{Visualization of Attention:} \\
    Advanced tools can visualize attention scores, accentuating the parts of the post that majorly influenced the model's conclusion. Words such as ``feeling", ``loneliness", and ``isolated" might be intensified, illustrating their central role in the 'Social Aspect' dimension classification.
\end{enumerate}

Understanding where the model is focusing allows insights into its decision-making process. This knowledge not only reveals model operations but can also aid users or developers to trust the model's results or pinpoint training needs.

\section{Experiments and Evaluation}
Extensive experiments for multi-class classification and explanation extraction were conducted to assess the effectiveness of the baselines and to identify their limitations.

\subsection{Baselines} 
In the multifaceted landscape of multi-class classification, a plethora of algorithms and methodologies have been devised and employed. Our study embarks on a detailed examination of four distinct types of multi-class classifiers, spanning both classic machine learning and contemporary deep learning models.

\subsubsection{Classic Machine Learning Algorithms}
The very foundation of our investigation stems from \textbf{classic machine learning algorithms}. Here, the focus is on employing traditional, yet powerful, classifiers. Our text data is initially transformed into a numerical representation through the Term Frequency-Inverse Document Frequency (TF-IDF) technique. This representation encapsulates the importance of words in relation to the entire corpus. Logistic Regression (LR) is a statistical model used for predicting the probability of an instance belonging to a particular category. Given its robustness and simplicity, LR is often a go-to for many text classification tasks. As an ensemble learning method, Random Forest (RF) constructs a multitude of decision trees during training and outputs the class that is the mode of the classes of the individual trees. Owing to its ability to mitigate overfitting and handle large datasets with higher dimensionality, it's a prominent choice for various classification challenges.

\subsubsection{Recurrent Models}
Delving into the realm of deep learning, we focus on \textbf{recurrent models}. Given the sequential nature of textual data, Recurrent Neural Networks (RNNs) are apt choices as they are innately suited to handle sequences. To represent the text, we employ pretrained word2vec embeddings. These embeddings capture semantic information and relationships between words. Within the recurrent paradigm, we deploy Long Short-Term Memory (LSTM) and Gated Recurrent Units (GRU). LSTM units are a type of RNN architecture. They possess the ability to remember patterns over long durations, thereby making them effective for sequence-based tasks. As per \cite{zulqarnain2020text}, GRUs are a variant of LSTMs. They come with a simpler structure, requiring fewer parameters and often achieving comparable performance. Both LSTM and GRU models in our study are constructed with two layers, comprising 32 hidden neurons each. The non-linear ReLU activation function is chosen to introduce non-linearity into the model. For the training process, we harness the power of the cross-entropy loss function, which serves as a good measure for classification tasks. Additionally, the Adam optimization algorithm, with a learning rate set at 0.001, is used to refine our model parameters and guide the training process to convergence.

%

\subsubsection{Encoder only Transformers}
Our investigation encompasses several variants of \textbf{encoder-only Transformers}. These models primarily focus on the encoder component of the original Transformer architecture. They are adept at capturing the context of every token in a sequence, making them particularly useful for tasks such as text classification, named entity recognition, and more. For our study, the input text is tokenized using a pre-trained Transformer tokenizer, resulting in 768-dimensional vectors. These vectors serve as the initial embeddings for our models. The models we consider are:

\begin{enumerate}
\item \textbf{BERT (Bidirectional Encoder Representations from Transformers)}: Introduced by \cite{kenton2019bert}, BERT is one of the pioneering encoder-only Transformer models. Its uncased version, BERT-base uncased, signifies that the model does not differentiate between upper and lower-case letters, treating the text in a case-agnostic manner.

\item \textbf{RoBERTa (A Robustly Optimized BERT Pretraining Approach)}: As an enhancement to BERT, RoBERTa-base~\cite{liu2019roberta} diverges from BERT in terms of training data and methodology, leading to more robust representations.

\item \textbf{ALBERT (A Lite BERT)}: ALBERT-base v2~\cite{lan2019albert} is a more efficient variant of BERT, offering similar or even better performance using significantly fewer parameters. It achieves this by factorizing the large matrix into smaller matrices, thereby reducing redundancy.

\item \textbf{DeBERTa}: An advanced variant of BERT, DeBERTa improves upon BERT by utilizing a disentangled attention mechanism, which allows different parts of the model to focus on different types of information.

\item \textbf{PsychBERT}: Specifically tailored for psychological text analysis, PsychBERT embodies domain-specific knowledge making it especially suitable for tasks in the realm of psychology.

\item \textbf{ClinicalBERT}: As the name suggests, ClinicalBERT is fine-tuned for clinical text, incorporating the nuances and lexicon commonly encountered in medical records and clinical narratives.

\item \textbf{MentalBERT}: With a focus on mental health texts, MentalBERT captures patterns and contexts pertinent to mental well-being, disorders, treatments, and other related facets.

\end{enumerate}

\subsubsection{Decoder-only Transformer}

On the other hand, the \textbf{decoder-only Transformer} primarily employs the decoder component of the Transformer architecture. Its strength lies in generating sequences, making it more apt for tasks like text generation, completion, and translation. Proposed by OpenAI in \cite{radford2018improving}, GPT stands as one of the flagship decoder-only Transformers. It's pre-trained on vast text corpora to generate coherent and contextually relevant sequences of text.

It should be noted that while encoder-only models like BERT are designed for tasks that require understanding of text context, decoder-only models like GPT are optimized for generating coherent sequences. However, the boundaries between these applications have been blurred in recent advancements, with models being adapted for a wider array of tasks than their initial design intention.


\textbf{Parameter Optimization}: For consistency, we used the same experimental settings for all models and used 10 fold cross-validation. All results are reported as the average across all folds. We used the grid search optimization technique to optimize the parameters. To tune the number of layers (n), we empirically experimented with the values: learning rate lr \(\in\) \{0.001, 0.005, 0.0001, 0.0005, 0.00001\} and optimization $O\ \in\ \{$Adam, Adamax, AdamW$\}$ with a batch-size of $\{8, 16, 32\}$. We used the base version pre-trained language models (LMs) via HuggingFace\footnote{https://huggingface.co/models}. 
We use optimized parameters for each baseline, and evaluate precision, recall, F1-score, and Accuracy. Varying lengths of posts are padded. We trained for 150 epochs with early stopping with a patience of 10 epochs. Thus, we set hyperparameter for our experiments with Transformer-based models as $H$ = $200$, $O$ = Adam, learning rate = 1$\times 10^{-5}$, batch size $16$, and $10$ epochs. We further regularize LSTM and GRU with kernel regularization and bias regularization of $1\times 10^{-4}$ learning rate.

\subsection{Quantitative Analysis}
Table~\ref{mainresults} displays the performance of different models on four wellness dimensions, namely Physical Aspect (PA), Social Aspect (SA), Emotional Aspect (EA), and Spiritual Aspect (SEA). From the table, it is evident that the GPT model and MentalBERT outperforms all the existing methods in terms of quantitative evaluation, indicating comparable results. While other models might have their respective strengths in various nuances, when it came to a holistic quantitative evaluation, these two models exhibited dominance. The GPT model and MentalBERT model leads the pack, eclipsing the performance metrics of all other existing methodologies. A plausible explanation lies in the lexical characteristics associated with each dimension. The Social Aspect often involves direct terms like family, breakup, friends, etc., which might be easier for models to recognize. On the other hand, the Spiritual Aspect's lexical universe, containing words like motivation, happy, sad, etc., is broader and possibly harder to pinpoint to a single aspect. 

\begin{table*}[ht]
\centering
\caption{Comparison of state-of-the-art methods. F-score, Precision, and Recall scores are averaged over 10 folds.}
\label{mainresults}
\begin{tabular}{l|ccc|ccc|ccc|ccc|c}
 \toprule[1.5pt]
\textbf{Method} & \multicolumn{3}{c}{\textbf{PA}}& \multicolumn{3}{c}{\textbf{IVA}}& \multicolumn{3}{c}{\textbf{SA}}& \multicolumn{3}{c}{\textbf{SEA}} & \textbf{Accuracy}\\ 
\hline
& P & R & F & P & R & F & P & R & F & P & R & F & \\
\midrule

\textbf{LR}& 0.68 & 0.51 & 0.58 & 0.78 & 0.38 & 0.51 & 0.53 & 0.96 & 0.69 & 0.70 & 0.32 & 0.44 & \textbf{0.6009} \\
\textbf{RF} & 0.45 & 0.47 & 0.46 & 0.44 & 0.36 & 0.40 & 0.49 & 0.63 & 0.55 & 0.42 & 0.26 & 0.32 & 0.4604\\
\midrule

\textbf{LSTM} & 0.57 & 0.60 & 0.59 & 0.71 & 0.57 & 0.63 & 0.72 & 0.82 & 0.77 & 0.57 & 0.50 & 0.54 & \textbf{0.6591} \\
\textbf{GRU} & 0.43 & 0.73 & 0.54 & 0.80 & 0.45 & 0.57 & 0.82 & 0.77 & 0.79 &  0.52 & 0.34 & 0.41 &  0.6316  \\



\midrule

\textbf{BERT} & 0.71 & 0.79 & \textbf{0.75} & 0.75   &   0.75 & 0.75  & 0.87 & 0.78 & 0.82 & 0.58 & 0.62 & 0.60 &  0.7464  \\
\textbf{RoBERTa} & 0.75 & 0.75 & \textbf{0.75} &  0.65  & 0.92 & 0.76 &  0.89 & 0.78  &  \textbf{0.83}  & 0.60 & 0.57  & 0.58  & 0.7480 \\
\textbf{ALBERT} & 0.75 &	0.73 &	0.74 &
0.70 &	0.77 &	0.73 &
0.82 &	0.83 &	0.82 &
0.61 &	0.57 &	0.59 & 0.7406 \\

\textbf{DeBERTa} & 0.80 &	0.76 &	0.78 &
0.82 &	0.76 & 	0.79 &
0.80 &	0.88 &	0.84 &
0.66 &	0.62 &	0.64 &\textbf{0.7779} \\

\midrule
\textbf{PsychBERT}  &	0.76	& 0.73 &	0.74 & 0.75 &	0.80 & 0.78  &	0.82	& 0.86 &	0.84 &	0.65 &	0.58 &	0.61& 0.7617\\
\textbf{MentalBERT} & 0.77	& 0.75	& 0.76 & 
0.79 &	0.81 &	0.80 &
0.83 &	0.88 &	0.85 &
0.68 &	0.61 &	0.64 & \textbf{0.7812} \\

\textbf{ClinicalBERT} & 0.74 &	0.77 &	0.75 &
0.73 &	0.75 &	0.74 &
0.83 &	0.86 & 0.84 &
0.68 &	0.57 &	0.62 & 0.7617\\

\midrule
\textbf{GPT-Ada} & 0.77 &	0.84 &	0.80 &
0.76 &	0.77 & 0.77 &
0.84 &	0.83 &	0.84 &
0.67 &	0.61 &	0.64 & 0.7779\\
\textbf{GPT-Babbage} & 0.80 &	0.75	& 0.77 &
0.82 &	0.83 &	0.83 &
0.79 &	0.85 &	0.82 &
0.69 &	0.64 &	0.66 & 0.7795 \\
\textbf{GPT-Curie} & 0.78 &	0.78 &	0.78 &
0.77 &	0.81 &	0.79 &
0.81 &	0.84 &	0.82 &
0.67	& 0.59 &	0.63 & 0.7698 \\
\textbf{GPT-3 Davinci} & 0.79 &	0.78 &	0.79 &
0.82 &	0.84 &	0.83 &
0.80 &	0.85 &	0.83 &
0.68 &	0.59 &	0.63 & \textbf{0.7812} \\
\midrule


 \bottomrule[1.5pt]
\end{tabular}
\end{table*}

The GPT model, built upon the Transformer architecture, is adept at capturing long-range contextual information in text. Its multiple attention mechanisms ensure that the model considers not just immediate neighboring words but also distant words to infer meaning. This becomes critical when determining context in intricate textual scenarios.
MentalBERT, on the other hand, benefits from a nuanced fine-tuning process. While it borrows from the BERT architecture, its specific training on mental and wellness-oriented datasets primes it for excellence in tasks involving psychological and wellness dimensions. Its specialization makes it particularly effective in this domain. An interesting facet of the evaluation was the comparability of GPT and MentalBERT's results. While both were at the top of the leaderboard, their performance metrics were notably close. This suggests that while GPT's generalized training makes it a powerhouse in various NLP tasks, MentalBERT's specialized training ensures it's not far behind, at least in this domain. The neck-to-neck performance of these models presents an intriguing proposition for researchers and practitioners. It raises questions about the merits of broad generalized training versus specialized domain-specific training in mental healthcare domain. To this end, we further examine the attention over the text-spans that is used for decision-making by these models, respectively.

The \textbf{Social Aspect (SA)} pertains to interactions and relationships one has with their peers, family, and the broader community. Our analysis discerns that SA exhibits a unique linguistic signature. This is manifested in the form of specific words and phrases that are intricately woven into the social tapestry of human interactions. For instance: Family: this word encapsulates the primary social unit, indicating close bonds, familial ties, and inherent responsibilities. Its presence can immediately point towards topics related to kinship or family dynamics. Contrastingly, the \textbf{Social-Emotional Aspect (SEA)} encompasses a broader spectrum of emotions and sentiments that bridge both social interactions and individual emotional states. This dimension's vocabulary, while relevant, tends to be more generic and widespread across various contexts. For example, Motivation: While crucial to understanding one's drive and determination, motivation is a multifaceted term that can span numerous topics, not strictly limited to social-emotional contexts. Thus, we observe the highest and lowest scores for SA and SEA, respectively. This could be attributed to the fact that SA has a specific set of words associated with it, such as \textit{family, breakup, friends, ditched, partying}, whereas SEA contains more general words, such as \textit{motivation, happy, sad, hate}. 

It was interesting to note that the identification of wellness dimensions is a challenging task, and the success of GPT can be attributed to its ability to interpret the contextual information present in social media text. However, there are some challenges that can make it difficult for pre-trained Transformers to accurately identify wellness dimensions in social media data. For example, social media data can be very informal and contain non-standard language, such as slang or abbreviations, which can be difficult for models to understand. Additionally, wellness dimensions can be highly subjective and context-dependent, which can make it challenging for models to accurately identify them. 
To ameliorate the issues posed by the idiosyncratic nature of social media text and the subjectivity in wellness dimensions, we advocate for a two-pronged approach: (i) Fine-tuning pre-trained transformer models on specific datasets, like ours, to make them more attuned to the linguistic nuances and context, and (ii) Investing in research to design models with a robust capability to process informal language and discern context, enabling them to perform even better in such tasks.
%




\subsubsection{Error Analysis}
Given that identifying wellness dimensions in social media involves analyzing text data in the context of psychological concepts, the task is highly complex and specific to the field of social computing. Consequently, the challenges associated with this task are further compounded by the need to incorporate domain-specific psychology into the analysis. With this in mind, we have identified the following NLP-centered challenges that must be addressed to ensure accurate and effective analysis of social media data for wellness dimensions:
\begin{enumerate}
    \item \textbf{Semantic Word Ambiguity}: Developing AI models for this task may result in semantic ambiguity during decision making. Consider a posts $P_2$ and $P_3$:
    \begin{quote}

        $P_2$: ...bad luck due to demons in my head...\\
        $P_3$: ...head-ache or injury in my head...
    \end{quote}
    Post $P_2$ affects \textsc{sea} in its discussion about omen, while $P_3$ pertains to \textsc{PA} through physical injury or incapability causing mental disturbance. Although, the word \textit{head} is used in both $P_2$ and $P_3$, the semantic interpretations are different.
    \item \textbf{Metaphors}: The cultural aspect of using metaphors is common practice in social media engagement~\cite{bail2016cultural,chmielecki2013conceptual}. Consider the following posts $P_1$ and $P_2$:
    \begin{quote}

        $P_1$: ...maybe I'll drink myself to death before I wind up homeless...\\
        $P_2$: ...I drink a lot of alcohol...
    \end{quote}
    Although both $P_1$ and $P_2$ contain the content word \textit{drink}, the expression \textit{drink myself to death} is a metaphor suggesting \textsc{iva} due to the financial risk of homelessness. On the contrary, $P_2$ points towards physical unfitness, i.e., \textsc{pa}.
    
    \item \textbf{Attention and Ambiguity}: A surge in discourse and pragmatics suggests natural language understanding and low-level analysis to reduce ambiguity by identifying words more important than others, even when the same words are less important in other posts. For instance, consider the following post section:
    \begin{quote}
    
        $P_0$: From dealing with the fallout of my ex, to stressors at work, nothing compares to true loss of my baby boy. To everyone feeling shitty this New Year’s Eve, you are not alone.\\
        $P_1$: My mom says I cant work and controls my life...\\
        $P_2$: ...soul sucking job...\\
        $P_3$: ...unable to connect with my soul...\\
        $P_4$: I have 0 friends who would talk to me outside of work... 
    \end{quote}
    According to post $P_0$, the author's true loss is the loss of an infant, affecting their \textsc{sa}. However, it also contains words such as \textit{feeling shitty} and \textit{work}, reflecting \textsc{sea} and \textsc{iva}, respectively. While words such as \textit{work} are present but should not be emphasized in $P_1$ and $P_4$, avoiding selection of \textit{work} as potential text span depicting \textsc{iva}. Instead, more attention is required on words such as \textit{mom} and \textit{friends} in $P_1$ and $P_4$, respectively, thereby assigning it the \textsc{sa} category. Similarly, a given word \textit{soul} must be emphasized as part of an adjective phrase in $P_2$ but as a noun in $P_3$, thereby identifying it as \textsc{iva} and \textsc{sea}, respectively. 
\end{enumerate}

\subsubsection{Experimental Inferences.}In our recent study, we observed that the baseline models, even those regarded as advanced, exhibited significant limitations, especially when considered for critical applications like mental health assessment. Our rigorous evaluations indicated that, contrary to popular belief, even the state-of-the-art Transformer models are far from perfect, often making errors that could be deemed as careless in high-stakes environments. A pertinent concern that arose was the issue of overfitting. When these models are trained and subsequently evaluated on in-domain data drawn from a singular distribution, they tend to become exceedingly attuned to the peculiarities of that data. As a result, their performance may degrade considerably when introduced to out-of-domain data or data that showcases different attributes. A prime example is how a model trained on data from one social media platform might flounder when faced with data from another platform. This could be attributed to a myriad of factors, ranging from differences in user health behaviors, variances in socio-demographic indicators, or even the unique linguistic idiosyncrasies that users on different platforms exhibit. These intricacies necessitate a deep domain knowledge, which our \textsc{WellXplain} dataset strives to encapsulate. Thus, while the promise of AI in mental health is undeniable, our findings reiterate the importance of exercising caution, especially when the stakes are high.

\subsubsection{Discussion.} Wellness dimensions, by their very nature, are intricately interwoven. They often don't manifest as distinct, separate entities, but rather as complex intersections of various facets of an individual's well-being. A significant chunk of the discrepancies is traceable to the overlap between the Physical Aspect (\textsc{pa}) and the combined Spiritual and Emotional Aspect (\textsc{sea}). The reason for this is straightforward: many activities or events impinge on multiple wellness dimensions simultaneously.
Taking the example of crying, it's an action that exacts a physical toll (physical fatigue, dehydration, etc.) even as it signals a deep emotional upheaval. Thus, determining whether to categorize it under \textsc{pa} or \textsc{sea} is intrinsically challenging. Social media posts, often reflective of real-life experiences, are rarely one-dimensional. A single post might touch upon various wellness dimensions, further complicating the annotation process. For instance, a post discussing a rigorous yoga session could touch upon the physical exertion involved (\textsc{pa}) and the ensuing mental tranquility (\textsc{sea}). The inherent subjectivity and variability of human emotions mean that many posts don't fit neatly into predefined categories. Ambiguity in textual content makes it hard to pin down the exact wellness dimension it pertains to. Reading between the lines, understanding subtext, or gauging the unsaid is a skill that varies among annotators. This variability is a fertile ground for manual disagreements.

\subsection{Qualitative Analysis}
To begin our investigation into monitoring the progression of mental disturbance, we concentrate on the wellness concept extraction through relevant text-spans. 
This preliminary study is a crucial step towards understanding the underlying mechanisms that lead to severe mental illnesses such as depression and self-harm by identifying wellness concept in a given text. Future work will aim to incorporate additional factors to gain a more comprehensive understanding positivity and negativity of wellness concepts in Reddit posts. In this section, we examine the reliability and explainability of the traditional multi-class classifiers.

\paragraph{Reliability/ Explainability Analysis.}
%

In our endeavor to assess the clarity and understandability of model predictions, we juxtaposed the explanations derived from ground truth with those garnered through the LIME method, as delineated in \cite{zirikly2022explaining}. Our experimental setup involved extracting pivotal terms from a sample of 100 randomly selected data points, employing both recurrent models and Transformer architectures. LIME's strength lies in its ability to identify and highlight those specific words within the data that most potently sway the classifier's verdict.

In our research, the analytical results, captured comprehensively in Table~\ref{tab:5}, incorporated two renowned metrics: ROUGE and BLEU. These metrics are universally recognized benchmarks in the domain of natural language processing, particularly when one seeks to evaluate the quality and relevance of generated text against a predetermined reference or gold standard. ROUGE, which stands for Recall-Oriented Understudy for Gisting Evaluation, primarily assesses the recall rate of generated content, spotlighting how many of the reference's components (words, phrases, etc.) were accurately captured in the produced text. On the other hand, BLEU (Bilingual Evaluation Understudy) emphasizes precision, ensuring that the generated content's components are indeed present in the reference text. Together, these metrics offer a holistic evaluation, gauging both the comprehensiveness and accuracy of generated explanations relative to a benchmark. Through the simultaneous application of ROUGE and BLEU, we ensured a thorough and robust assessment of the explanations' fidelity to the reference standard.

Our experimental outcomes presented insightful revelations about the performance of various language models. Notably, GPT-3 ascended as the frontrunner, demonstrating unparalleled precision, recall, and F-score metrics, illustrating its prowess in accurately identifying and spotlighting critical words within generated sequences (See table~\ref{tab:5}). Following closely was the performance exhibited by the MentalBERT model. Unlike GPT-3, which gauges impactful words through its generated sequences, MentalBERT employs the LIME methodology, with a specific emphasis on the attention mechanism to earmark salient words.


\begin{table*}
    \centering
\caption{LIME's explanation scores using ROUGE-1.}
    \begin{tabular}{l|ccc|cc}
        \toprule[1.5pt]
        \textbf{Method} & \textbf{F-score} & \textbf{Precision} & \textbf{Recall} & \textbf{Bleu-1} & \textbf{Bleu-2}\\
        \midrule 
\textbf{BERT} & 0.4668 & 0.3970 & 	0.7076 &		0.3687 &	0.1311 \\
\textbf{RoBERTa} & 0.4170  & 0.3197& 0.8525  & 0.3651 & 0.1302 \\
\textbf{ALBERT} &		
0.4660 &	0.3954 &	0.6999 &		0.3682 &	0.1342 \\

\textbf{DeBERTa} &
0.4564 &	0.3886 &	0.6852 &		0.3614 & 0.1342 \\
	\midrule				
\textbf{PsychBERT} & 0.4581 &	0.3890 &	0.6879 &	0.3647 &	0.1383 \\
\textbf{MentalBERT}		&			
0.4866 &	0.4095 &	0.7463 &		0.3827 &	0.1456 \\
\textbf{ClinicalBERT} &				
0.4691 &	0.3985 &	0.7105 &		0.3720 &	0.1401 \\
\midrule
\textbf{GPT-Ada} &	0.5002 &	0.5548 &	0.5543 &	0.4078 &	0.3485 \\
			\textbf{GPT-Babbage} &		
0.5267 &	0.5944 &	0.5761 &	0.4226 &	0.3591 \\
		\textbf{GPT-Curie} &				
0.5213 &	0.5874 &	0.5722 &		0.4240 &	0.3582 \\
			\textbf{GPT-Davinci}	&		
0.5335 &	0.5930 &	0.5916 &		0.4361 &	0.3758 \\
									        
        \bottomrule[1.5pt]
    \end{tabular}
    
    \label{tab:5}
\end{table*}

It's pertinent to mention that while GPT-3 and MentalBERT both achieved commendable results, drawing a direct comparison between the two may not be straightforward, given their differing methodologies and underlying architectures. Notwithstanding, the distinctions in their results were conspicuous, thereby underscoring the respective strengths and nuances of each approach.

A key observation that further emerged from our study was the superior reliability offered by expansive generative language models like GPT-3. Despite the tailor-made fine-tuning that models like MentalBERT undergo for domain-specific tasks, the expansive training and inherent capabilities of models like GPT-3 enable them to outshine in various tasks. This observation potentially accentuates the evolving dynamics of AI models, wherein broad-spectrum models, thanks to their extensive training and diversified datasets, might exhibit more robust and reliable performances, even when juxtaposed against domain-specialized transformers.

\paragraph{Practical Implications.}
We suggest longitudinal studies for fair and accountable practices of analyzing the emotional spectrum of users' historical social media posts~\cite{sawhney2021towards}. The ongoing research aimed at identifying changes in moments from users' longitudinal social media posts~\cite{tsakalidis2022overview} should benefit from our \textsc{Wellness Dimensions} dataset. Consider the following set of posts posted by a user $A$ for varying time intervals $t=\{t1, t2, t3, t4, t5\}$: 

\begin{quote}
\emph{T1:} I am not entirely sure; I am making sense of my life. \\
\emph{T2:} Politics is disastrous, and I am in the middle. \\
\emph{T3:} It drives me on the edge and restless to see what the outcome would be. \\
\emph{T4:} I hope there would be a life without politics. My relationship with my wife is also political. \\
\emph{T5:} I need better life after my death.
\end{quote}

Post $T1$ illustrates the confused state of the social media user. Movement of thoughts towards politics $(T2 \rightarrow T3)$, relationships ($T4$) and finally towards suicidal intention ($T5$) illustrates the tremendous impact of users' experiences on different wellness dimensions over time. Henceforth, our dataset shall support and compliment other studies~\cite{saha2022towards,gaur2021characterization} as an intrinsic classification task.

\paragraph{Limitations and Ethical Consideration}
We acknowledge that our work is subjective in nature and thus, interpretation about wellness dimensions in a given post may vary from person to person. 
Clearly, machine learning predictions are unable to replace professional mental health diagnostic let alone counseling and therapy. As shown in our evaluation, their accuracy and trustworthiness remain insufficient for such purposes~\cite{nicholas2020ethics}. Rather, we are hoping to engender further research on how to recognize early warning signs that may otherwise go unnoticed and neglected. For this, it is important to obtain consent for all relevant use cases. 
It is also important for human professionals to carefully assess model predictions before any pertinent action is taken. Our work promotes explainability to facilitate human validation~\cite{chancellor2019taxonomy}. However, even here it must be noted that explanations can be misleading and that human validators need to very carefully review the entire post context. 

We emphasize the importance of preserving privacy due to the sensitive nature of social media data~\cite{harrigian2021state,chancellor2019taxonomy}. To ensure our accountability, we will provide appropriate protections for sensitive data, and there
will be no linkage of the dataset to other sites that
could jeopardize user anonymity~\cite{shing2018expert}. We further acknowledge that suggesting professional help based on a given social media post is a starting point for this study. In our work, \emph{explainability} refers to the text spans that appears as indicators of wellness dimensions in a given text.
    
      
         
    

\section{Conclusion}
In this work, we present \textsc{WellXplain}, a newly constructed dataset for wellness concept extraction and classification to facilitate the future research in this direction. This work is based on a new task definition of clinical concept extraction and classification with a carefully designed annotation scheme, including perplexity guidelines. Furthermore, we solicit class annotations and text spans for explainability purposes. In addition, we present baseline experiments conducted on a diverse range of methods. 
The contribution of this work derives from the potential for tackling different use cases of wellness dimensions at a deeper, interpretable level. We endeavor to disseminate this position widely in the research community and urge researchers to develop richer, explainable models for inferring mental illness on social media. We keep data augmentation and few-shot learning approaches as open research directions to develop efficient AI models.
With this work, we seek to promote investigating mental health diagnostics that operate at a deeper, interpretable level and hope that future work benefits from our data. Furthermore, by considering the different dimensions of wellness, Social Determinants of Health 2030 can create policies that address the various aspects of well-being, leading to a more holistic evaluation~\cite{gomez2021practice}.

\section*{Acknowledgements}
We would like to convey our heartfelt thanks to our postgraduate student annotators, Ritika Bhardwaj, Astha Jain, and Amrit Chadha, for their meticulous contributions to the annotation process. Our deep gratitude goes to Veena Krishnan, an esteemed clinical psychologist, and Ruchi Joshi, a dedicated rehabilitation counselor, for their steadfast support and invaluable insights throughout the duration of this project. Furthermore, our warm appreciation is directed towards Prof. Sunghwan Sohn for his unwavering mentorship and guidance throughout our journey.

\bibliographystyle{elsarticle-num}

\bibliography{cas-refs}
\section*{Appendix}
\appendix

\section{Word Frequency in Explanations}
\label{wordcloud}
\begin{figure*}[htp]
    \includegraphics[width=.45\textwidth]{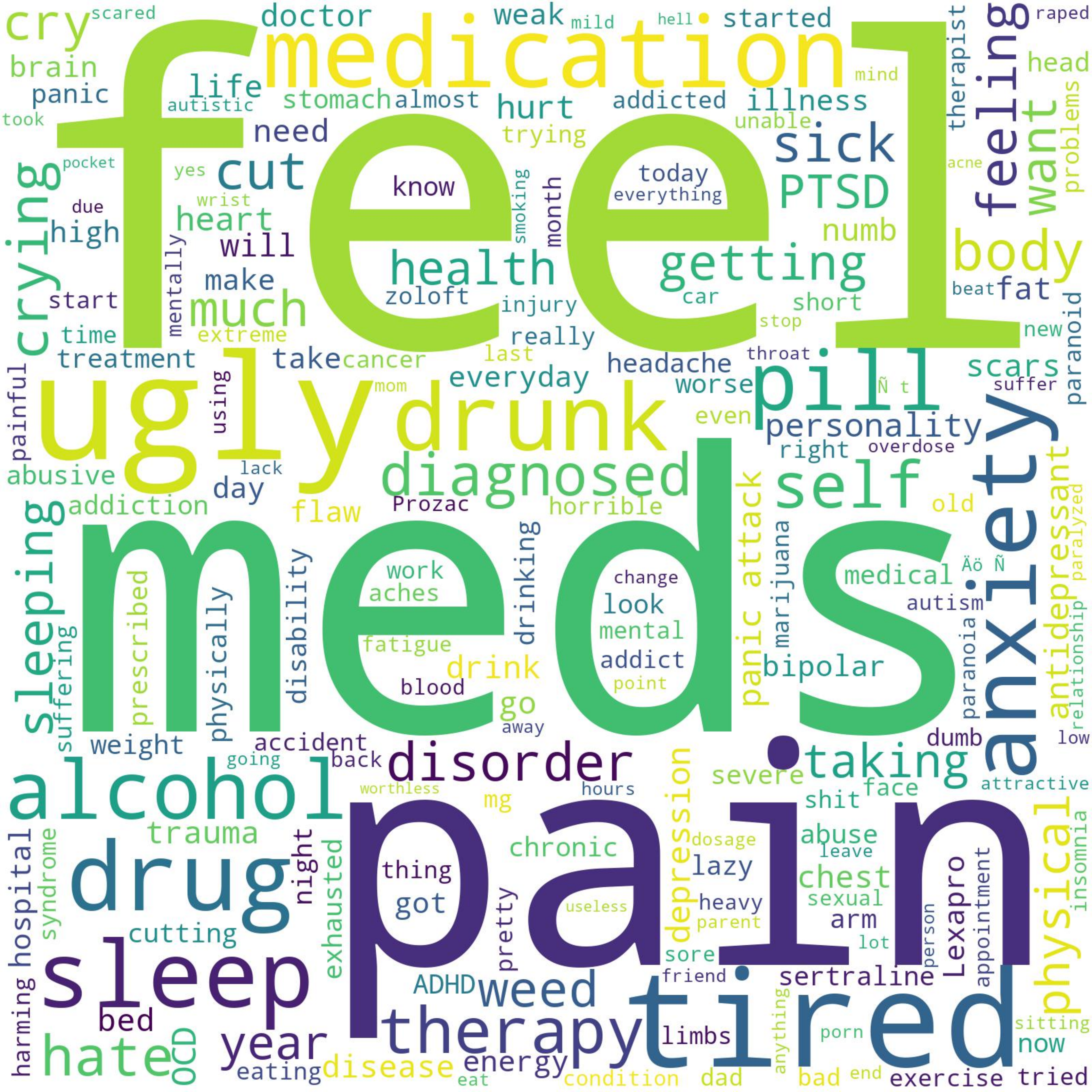}\hfill
    \includegraphics[width=.45\textwidth]{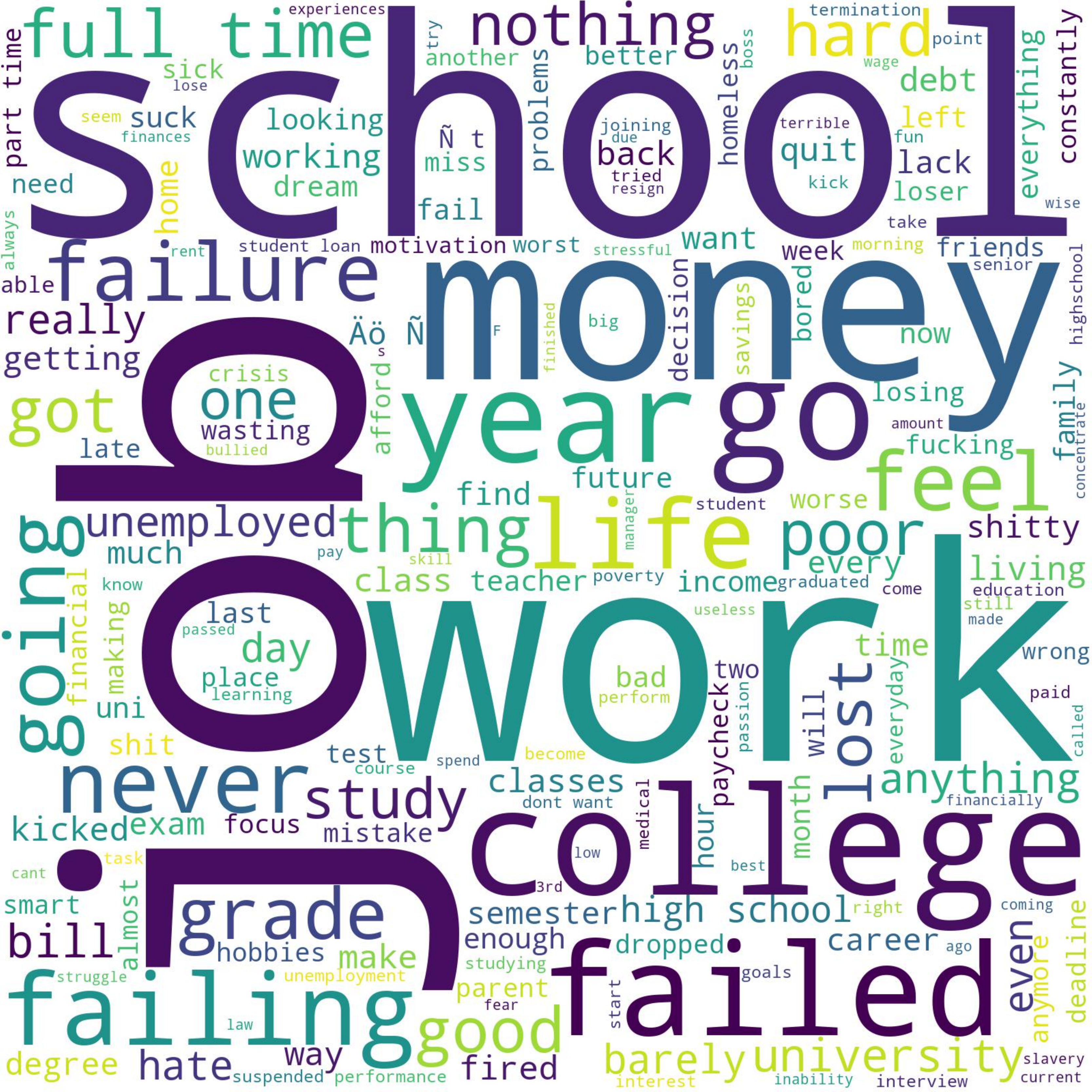}
    \\[\smallskipamount]
    \includegraphics[width=.45\textwidth]{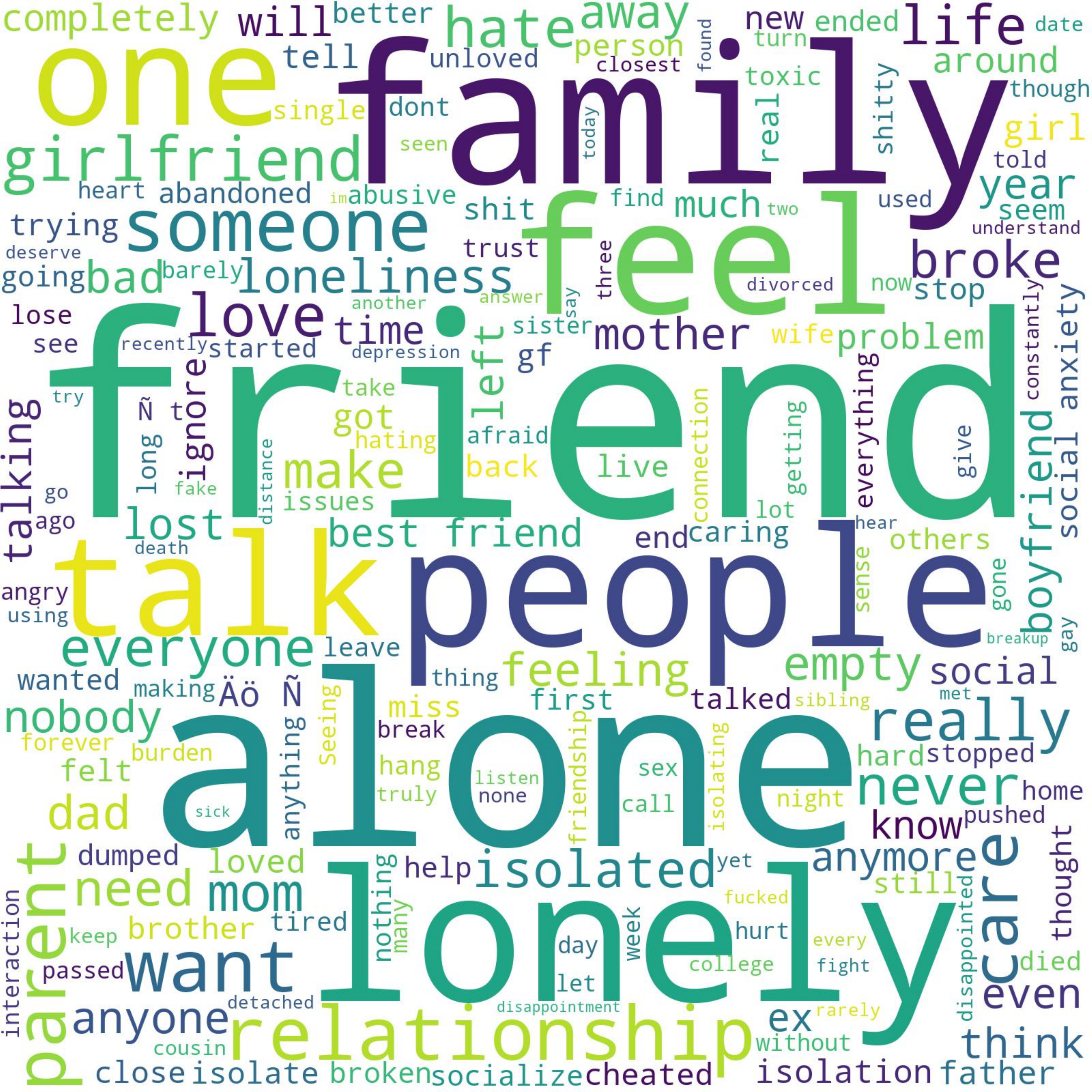}\hfill
    \includegraphics[width=.45\textwidth]{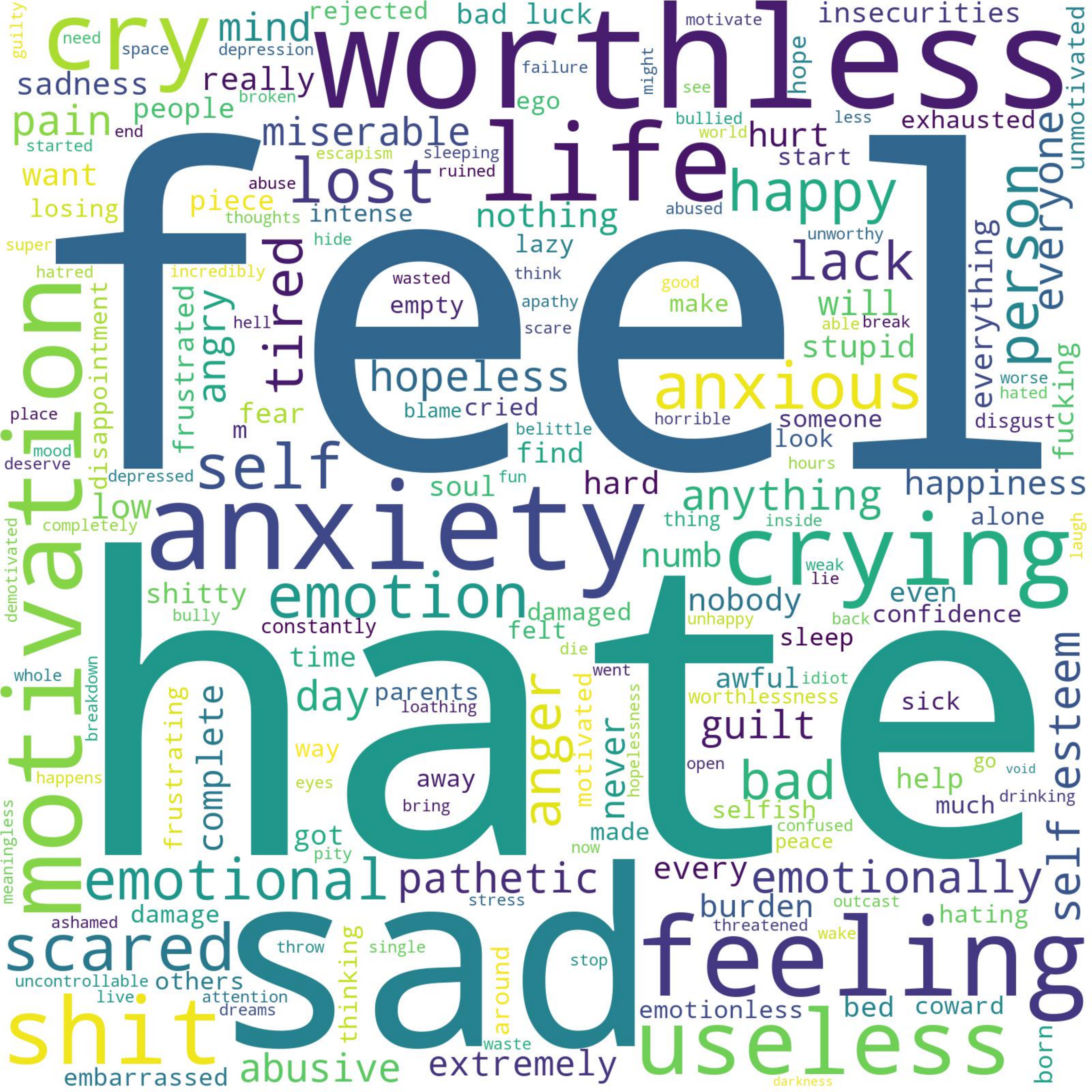}
     
    \caption{\textbf{Top}: Word cloud for Physical Aspect (Left) and  Intellectual and Vocational Aspect (Right);// \textbf{Bottom}: Word cloud for Social Aspect (Left) and Spiritual and Emotional Aspect (Right) }
    \label{fig:8}
\end{figure*}

Word clouds for explanations of different wellness dimensions are shown in Figure~\ref{fig:8}.





\end{document}